\documentclass{article}
\usepackage{spconf, amsmath, amsfonts, graphicx, svg, hyperref, mathtools, breqn, svg, caption, subcaption, tabularx}

\hypersetup{
    colorlinks=true,
    linkcolor=blue,
    filecolor=magenta,      
    urlcolor=cyan,
    pdftitle={Overleaf Example},
    pdfpagemode=FullScreen,
}
\newcolumntype{Y}{>{\centering\arraybackslash}X}

\title{Unsupervised Multiple Domain Translation through Controlled Disentanglement in Variational Autoencoder}

\name{Antonio Almudévar $^1$ \thanks{This work was supported by MCIN/AEI/10.13039/501100011033 under Grants PDC2021-120846-C41 \& PID2021-126061OB-C44, and in part by the Government of Aragón (Grant Group T36 23R). This project has received funding from the European Union’s Horizon 2020 research and innovation programme under the Marie Skłodowska-Curie grant agreement No 101007666. This work was performed using HPC resources from GENCI–IDRIS (Grant 2022-AD011012565). The research reported here was conducted at the 2023 Frederick Jelinek Memorial Summer Workshop on Speech and Language Technologies, hosted at Le Mans University (France) and sponsored by Johns Hopkins University.}, Théo Mariotte $^2$, Alfonso Ortega $^1$, Marie Tahon $^2$}
\address{$^1$ViVoLab, Arag\'{o}n Institute for Engineering Research (I3A), University of Zaragoza, Spain\\ 
$^2$LIUM, Le Mans Université, France\\
almudevar@unizar.es}

\begin{document}
\ninept
\maketitle
\begin{abstract}
Unsupervised Multiple Domain Translation is the task of transforming data from one domain to other domains without having paired data to train the systems. Typically, methods based on Generative Adversarial Networks (GANs) are used to address this task. However, our proposal exclusively relies on a modified version of a Variational Autoencoder. This modification consists of the use of two latent variables disentangled in a controlled way by design. One of this latent variables is imposed to depend exclusively on the domain, while the other one must depend on the rest of the variability factors of the data. Additionally, the conditions imposed over the domain latent variable allow for better control and understanding of the latent space. We empirically demonstrate that our approach works on different vision datasets improving the performance of other well known methods. Finally, we prove that, indeed, one of the latent variables stores all the information related to the domain and the other one hardly contains any domain information.
\end{abstract}
%
%
\vspace{-1mm}
\section{Introduction}
\label{sec:intro}
Domain translation involves the process of converting data from one domain to another while preserving the underlying information or structure. Multiple domain translation is a generalization of the previous task that implies translating between more than two domains simultaneously. In addition, when we do not use data pairs in the source and target domains to design and train the systems, they are said to be unsupervised. Typically, models based on Generative Adversarial Networks (GAN) \cite{goodfellow2014generative} are used to solve the domain translation problem \cite{mathieu2016disentangling, pix2pix2017, CycleGAN2017, yi2017dualgan, UNIT2017, MUNIT2018, gonzalez2018image, StarGAN2018, hui2018unsupervised, StarGANv22020, pang2021image}. However, GANs are usually complex to train and tune. Moreover, they present potential problems such as mode collapse \cite{goodfellow2014generative}. In this paper we present a method to solve the unsupervised multiple domain translation task using only a Variational Autoencoder (VAE) based model \cite{VAE2013}. Thus, we avoid the problems derived from using GANs. To the best of our knowledge, there are no other works using only a VAE to solve this task. 

\begin{figure}[hbt!]
    \centering
    \includegraphics[width=0.65\columnwidth]{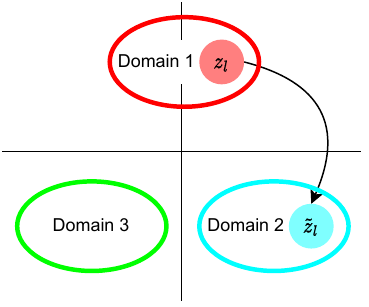}
    \caption{Scheme of how the translation is performed in a two-dimensional latent space with three domains. In this case, the input belongs to domain 1 and is translated to domain 2 by rotating the latent variable the same angle that separates the regions of the prior distributions of these two domains.}
    \label{fig:rotation_latent}
\end{figure}

Specifically, in this paper, we propose to use a VAE that has two independent latent spaces. One of them should contain information only about the domain, while the other one should contain information about everything that is not the domain, which we call style. Equivalently, this means that we try to disentangle the information corresponding to the domain and the style. Thus, to modify the domain and keep only the style, we will modify the latent space corresponding to the domain and leave the one corresponding to the style intact. To achieve this, we impose during training that the latent distribution corresponding to the domain must be close to a target distribution that depends on the domain. Specifically, we locate these distributions in different regions of the space according to the domain. In this way, we perform a transformation of the latent variable by bringing it to the region of the space corresponding to the target domain. In this work, we impose for efficiency that these transformations are linear and, more specifically, rotations. Thus, to translate from one domain to another we perform a rotation in the latent space, as schematized in figure \ref{fig:rotation_latent}, and then decode once this rotation has been performed. The fact of imposing that these transformations are linear also allows us to have a better control and understanding of the latent space. Other transformations could be explored, but this is beyond the scope of this paper. 

We could summarize the main contributions of this work in the following points:
\begin{itemize}
    \item We design a VAE in which the target distribution depends on the domain, but the encoder and decoder are common for every domain.
    \item We construct a VAE with two disentangled latent variables in which one of them depends only on the domain. 
    \item The domain translation is performed by means of a linear transformation latent variable corresponding to the domain.
\end{itemize}

The code to implement and perform the experiments explained in this paper is available at
\url{https://github.com/antonioalmudevar/variational_translation}.

\section{Related Work}
\label{sec:related_work}
\textbf{Domain Translation.} Different GAN-based works have achieved excellent results for domain translation. For example, pix2pix \cite{pix2pix2017} uses Conditional GANs \cite{cgan2014} and combines adversarial and L1 losses to train them. However, this method requires paired datasets. Other known methods are unsupervised, i.e., they do not require paired datasets. One of these is CycleGAN \cite{CycleGAN2017}, which introduces a cycle-consistency loss, which enforces that translated images can be converted back to their original domain. On the other hand, UNIT \cite{UNIT2017} makes a shared-latent space assumption and proposes an unsupervised image-to-image translation framework based on Coupled GANs. MUNIT \cite{MUNIT2018} extends the capabilities of UNIT by introducing separate content and style latent spaces for each domain, allowing for fine-grained control over content and style. The above proposals, while providing impressive results, only allow translation between two domains and not from multiple domains to multiple domains. The main method in the literature to solve this limitation is StarGAN \cite{StarGAN2018}. This uses a single generator and discriminator to translate between multiple domains. During training, the authors randomly generate a target domain label and train the model to flexibly translate an input image into the target domain. StarGANv2 \cite{StarGANv22020} builds upon StarGAN by introducing a more advanced architecture and improved disentanglement of content and style.

\vspace{1.5mm}
\noindent\textbf{Disentanglement in VAEs.} Disentangled VAEs aim to learn representations of data where different factors of variation are separated and interpretable. To achieve this, different approaches are relevant. $\beta$-VAE \cite{beta-vae2017} introduces a hyperparameter, $\beta$, that balances the reconstruction loss and a KL loss, which encourages the model to learn more disentangled representations. Additionally, Factor-VAE \cite{factor-vae2018} extends $\beta$-VAE by introducing a total correlation term in the loss function. On the other side, InfoVAE \cite{infovae2018} introduces an information-theoretic framework to disentangle factors of variation in data. Concretely, it uses mutual information between latent variables and data to encourage meaningful disentanglement. Finally, in \cite{ding2020guided}, they use classifiers that take the different latent variables as input, thus making each of this variables focus on an attribute. While all these approaches successfully disentangle various sources of variability, they do not provide the ability to specify the characteristics we wish to separate in advance. Determining which dimensions correspond to specific factors must be conducted after training these models \cite{kingma2014semi, kim2018disentangling, antoran2019disentangling}.

\vspace{1.5mm}
\noindent\textbf{Conditional VAEs.} In \cite{sohn2015learning} the Conditional VAE (CVAE) is presented, which is a variant of VAE in which the encoder, decoder, and prior distribution of the latent space depend on a condition. In this work, the authors suggest two alternatives: one is to make use of a neural network that takes the condition as input to define the prior distribution. The other is to simplify the problem by maintaining a consistent prior distribution regardless of the condition. In the present work, as we explain below, we make the encoder and the decoder do not depend on the condition . In addition, we define the condition-dependent prior without using a neural network, which allows us to perform domain translation in a very efficient way.

\section{Proposed Approach}
\label{sec:proposed_approach}

\begin{figure*}[t]
    \centering
    \includegraphics[width=0.75\textwidth]{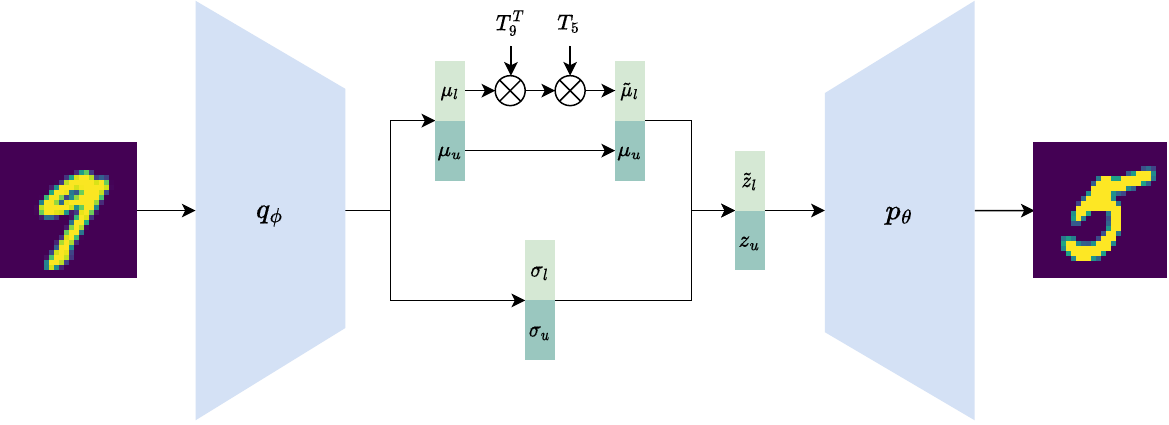}
    \caption{Scheme of how an input image $x$ whose class is $c$ is translated to class $t$. The original input is first encoded obtaining $p(z_l)$ and $q_\phi(z_u)$. Then, the mean of $q_\phi(z_l)$ is multiplied by $T_c^T$ and $T_t$ to obtain the mean of $q_\phi(\tilde{z}_l)$. Finally, a sample from this last distribution and another from $q_\phi(z_u)$ are decoded to obtain the translated version of the input. In this figure the input class $c$ is 9 and the target class $t$ is 5.}
    \label{fig:vae_scheme}
\end{figure*}

\subsection{Variational Autoencoder}
In the classical version of the Variational Autoencoder \cite{VAE2013}, the aim is to generate new data that closely resembles the input data from a dataset. To achieve this goal,  a latent variable $z$ is used, which can be considered to contain relevant information from the input $x$. To facilitate this process, authors propose the use of an encoder $q_\phi(z|x)$, a decoder $p_\theta(x|z)$, and a prior $p_\theta(z)$. This prior is usually defined as $\mathcal{N}(0, I)$, which implies that $p_\theta(z)$ is equivalent to $p(z)$. The ELBO to maximize in this case is:
\begin{equation}
    \mathcal{L}_{VAE} = \mathbb{E}_{q_\phi(z|x)}\left[ p_\theta(x|z) \right] - D_{KL} \left(q_\phi(z|x) || p(z) \right)
\end{equation}
Once the model has been trained to maximize this ELBO, one or more samples from $p(z)$ are drawn and passed through the decoder to obtain $p_\theta(x|z)$, enabling the generation of new samples.

\subsection{Conditional Variational Autoencoder}
In \cite{sohn2015learning}, the authors suggest conditioning data generation on specific conditions denoted as $c$. Typically, these conditions align with the class of the input $x$ in a labeled dataset. In this case we have an encoder $q_\phi(z|x,c)$, a decoder $p_\theta(x|z,c)$ and a prior $p_\theta(z|c)$. In the original proposal, this prior is proposed to be a neural network that takes the conditions $c$ as input. Different architectures for this networks can be used \cite{zhao2017learning}. However, in the majority of applications the relaxation $p_\theta(z|c)=p(z)=\mathcal{N}(0,I)$ is used \cite{lim2018molecular, pol2019anomaly, petrovich2021action}. The ELBO in the CVAE is:
\begin{equation}
    \mathcal{L}_{CVAE} = \mathbb{E}_{q_\phi(z|x,c)}\left[ p_\theta(x|z,c) \right] - D_{KL} \left( q_\phi(z|x,c) || p_\theta(z|c) \right)
\end{equation}

\subsection{Conditioning only the Prior Distribution}
One of the novelties we propose in this work is to condition the prior distribution, but not the encoder and decoder. Moreover, instead of using a neural network to condition this prior, we propose to define it by imposing conditions that will allow us to perform domain translation, as explained in the next two sections. In this case, therefore, we have an encoder $q_\phi(z|x)$, a decoder $p_\theta(x|z)$ and a prior $p(z|c)$ and the ELBO is:
\begin{equation}
    \mathcal{L}_{CPVAE} = \mathbb{E}_{q_\phi(z|x)}\left[ p_\theta(x|z) \right] - D_{KL} \left(q_\phi(z|x) || p(z|c) \right)
\end{equation}

\subsection{Disentanglement of Labeled and Unlabeled information}
The second fundamental novelty that we incorporate in this work is to include two conditionally independent latent variables with respect to the class $c$. One of those, denoted $z_l$, models the information corresponding to the known information depends on the class. The other one corresponds to all the information that is unlabeled and we denote $z_u$. Moreover, since they are conditionally independent, we have that $p(z_l,z_u|c) = p(z_l|c) p(z_u|c)$. Also, having imposed that $z_u$ does not depend on $c$, then it remains that $p(z_l,z_u|c) = p(z_l|c) p(z_u)$. Finally, we impose that $q_\phi(z_l,z_u|x)=q_\phi(z_l|x) q_\phi(z_u|x)$. Under these conditions, we are left with ELBO being:
\begin{align}
\begin{split}
    \mathcal{L}_{CDVAE} 
    & = \mathbb{E}_{q_\phi(z_l, z_u|x)}\left[ p_\theta(x|z_l, z_u) \right] \\ 
    & - D_{KL} \left(q_\phi(z_l|x) || p(z_l|c) \right)\\ 
    & - D_{KL} \left(q_\phi(z_u|x) || p(z_u) \right)
\end{split}
\end{align}
It's worth noting that during training, we maximize a modified version of the ELBO, in which both KL divergence terms are multiplied by a $\beta$ factor. In addition, multiple labeled latent variables could be defined, each conditionally independent of the others, in order to modify multiple input attributes. However, this is beyond the scope of the present work. 

Following the prevalent approach in VAEs, we define $q_\phi(z_l|x)=\mathcal{N}(\mu^l_\phi, \text{diag}(\sigma^l_\phi))$ and $q_\phi(z_u|x)=\mathcal{N}(\mu^u_\phi, \text{diag}(\sigma^u_\phi))$. This formulation ensures that $z_l$ and $z_u$ exhibit conditional independence with respect to $x$.

\vspace{-2mm}
\subsection{Defining prior distributions}
The last novelty is the way we define the priors, explained in the previous section. Defining these priors properly will allow us to perform domain translation. To do so, the first step is to define a vector $\mu_0$. We do this randomly so that each of its elements follows a $\mathcal{U}(0,1)$ distribution. Subsequently, we define a rotation matrix $T_c$ for each class $c=1,2,\dots,C$. Each of these rotation matrices $T_c$ is the normalized version of the Q matrix of a QR factorization of a random matrix in which each element follows a uniform distribution $\mathcal{U}(0,1)$. Finally, we define $\mu_c = T_c \cdot \mu_0$, $p(z_l|c)=\mathcal{N}(\mu_c, I)$, $c=1,2,\dots,C$ and $p(z_u)=\mathcal{N}(0,I)$. We should note that $\mu_0 = T_c^{-1} \cdot \mu_c = T_c^T \cdot \mu_c$, since $T_c$ is a rotation matrix and therefore orthogonal.

\subsection{Translating the domain}
\begin{figure}[hbt!]
    \centering
    \includegraphics[width=0.6\columnwidth]{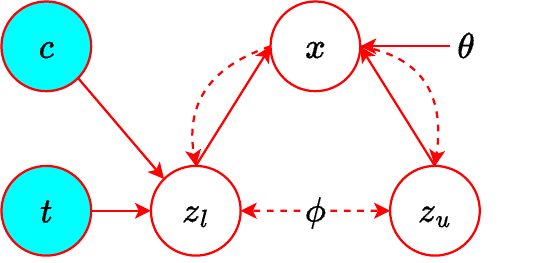}
    \caption{Graphical model of the translation method. Dashed lines correspond to the encoding processes $q_\phi(z_l|x)$ and $q_\phi(z_u|x)$. Solid lines correspond to the decoding process in which the target class $t$ is used to modify the latent variable $z_l$ before getting $p_\theta(x|z_l, z_u)$.}
    \label{fig:gm}
\end{figure}

Once the model is trained, we can perform the domain translation. For this, we have an input $x$, its class $c$, and a target class $t$. First, we obtain $z_l \sim q_\phi(z_l|x)$ and $z_u \sim q_\phi(z_u|x)$. Subsequently, we obtain the rotated version of $z_l$ as $\tilde{z}_l=T_t \cdot \tilde{z}_0$, where $\tilde{z}_0 = T_c^T \cdot z_l$. Finally, we obtain the translated version of the input as $\tilde{x} \sim p_\theta(x|\tilde{z}_l, z_u)$. We show the scheme of the proposed method and the graphical model of this domain translation in figures \ref{fig:vae_scheme} and \ref{fig:gm}, respectively.

\section{Results}
\label{sec:results}
\begin{figure*}[hbt!]
     \centering
     \begin{subfigure}[b]{0.49\textwidth}
         \centering
         \includegraphics[width=\textwidth]{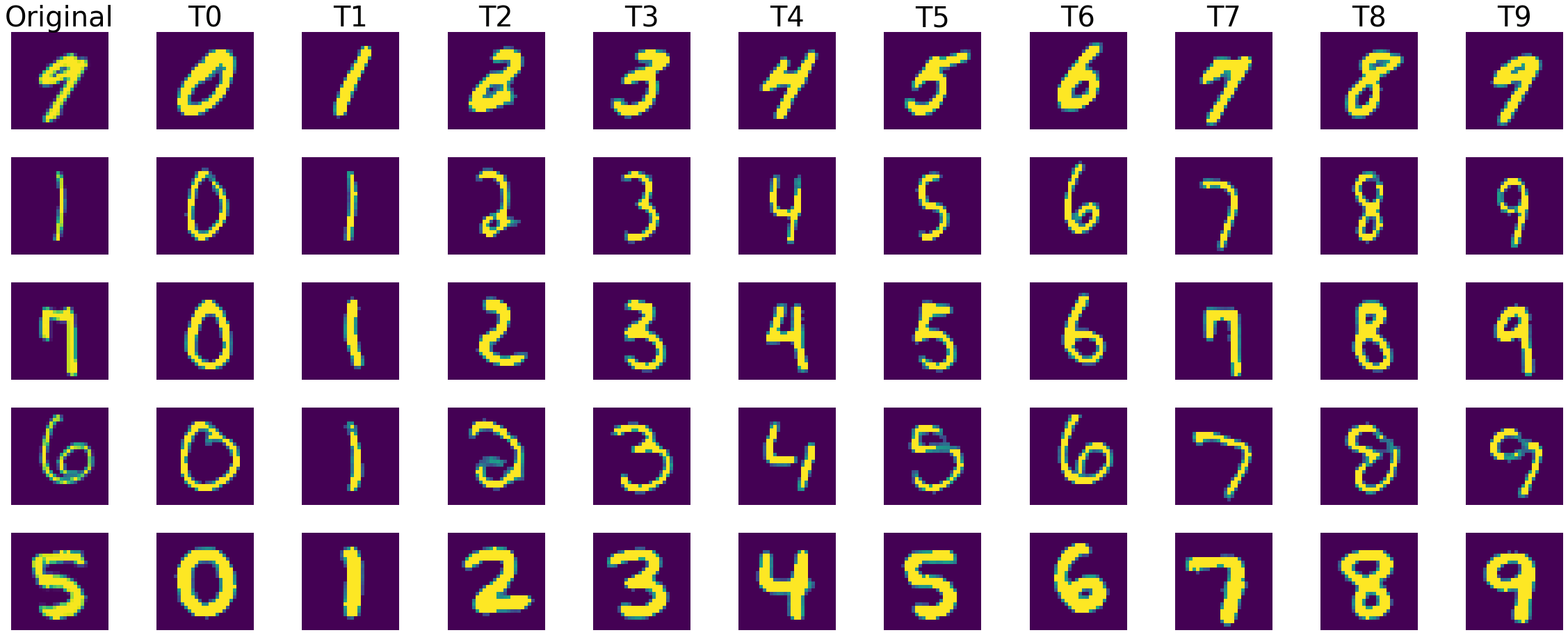}
         \caption{Ours}
     \end{subfigure}
     \hfill
     \begin{subfigure}[b]{0.49\textwidth}
         \centering
         \includegraphics[width=\textwidth]{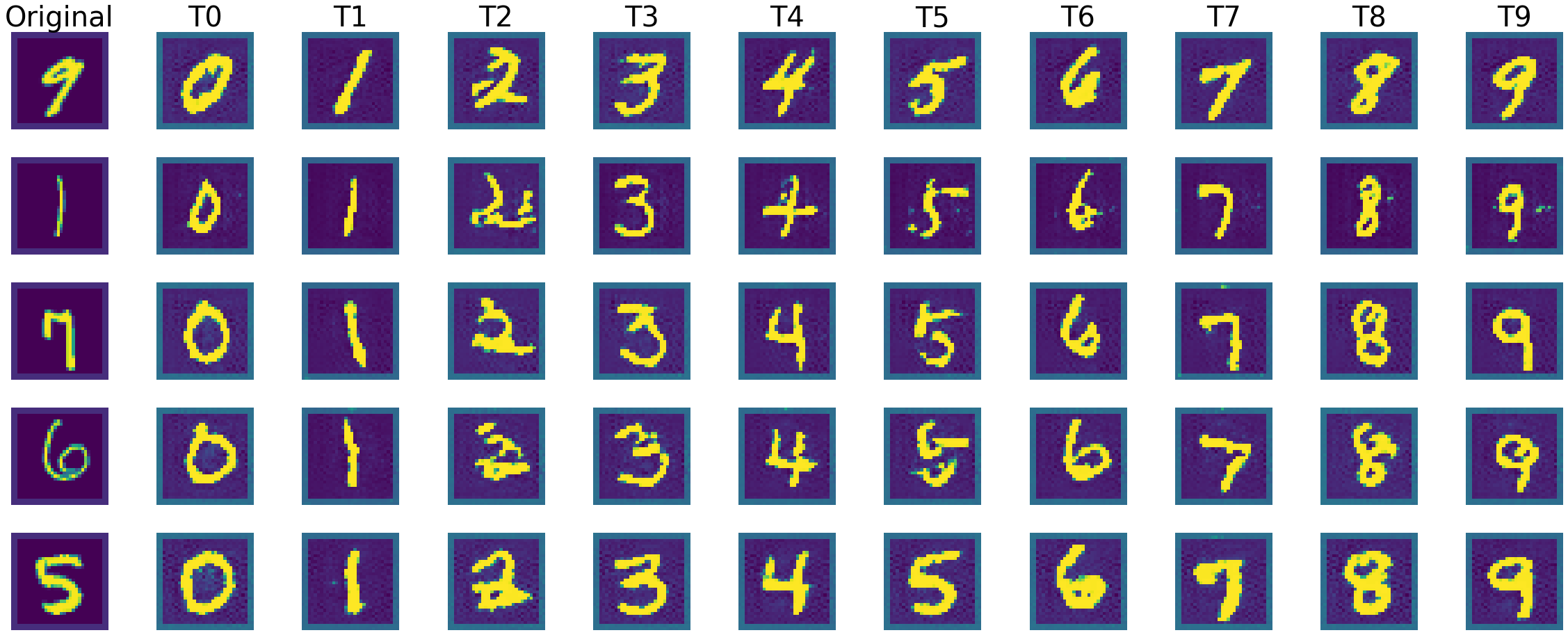}
         \caption{StarGAN \cite{StarGAN2018}}
     \end{subfigure}
    \caption{Comparison between our method and StarGAN for MNIST}
    \label{fig:mnist}
\end{figure*}

\begin{figure}[hbt!]
    \centering
    \includegraphics[width=0.49\textwidth]{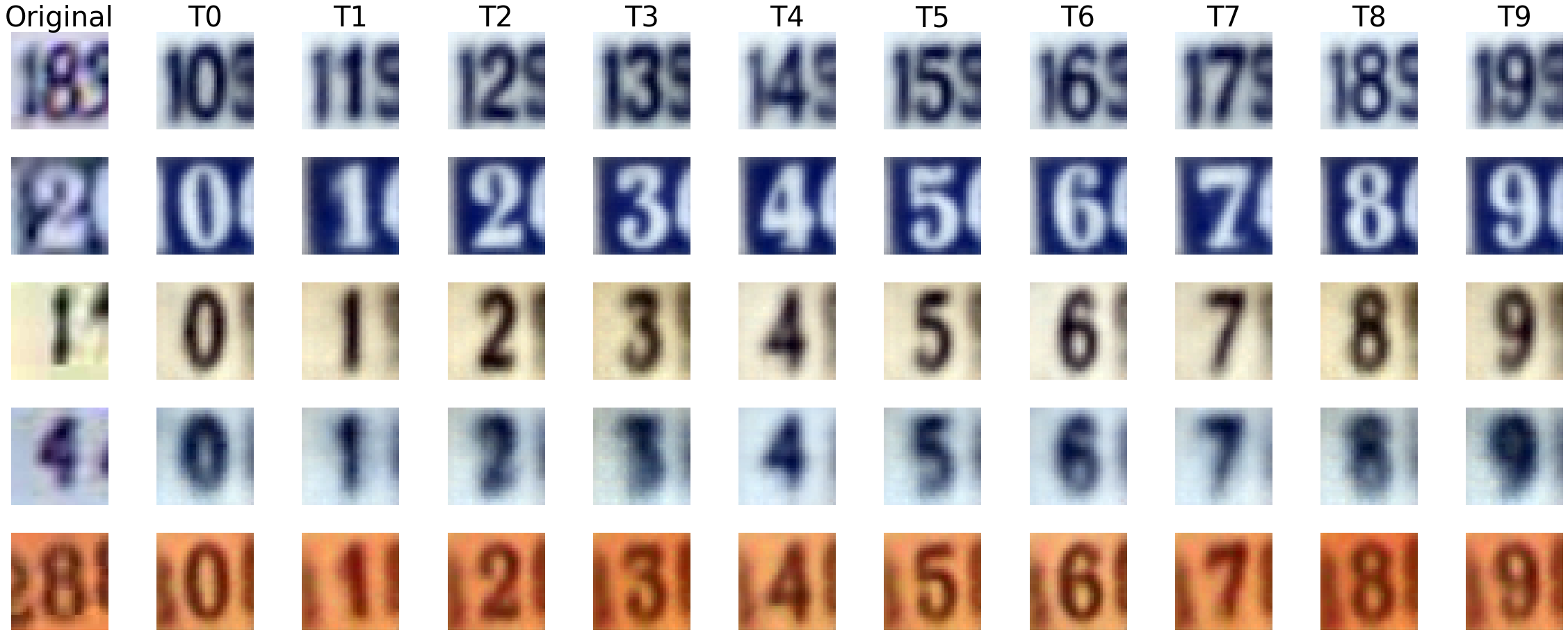}
    \caption{Results for SVHN}
    \label{fig:svhn}
\end{figure}

\begin{figure}[hbt!]
    \centering
    \includegraphics[width=0.49\textwidth]{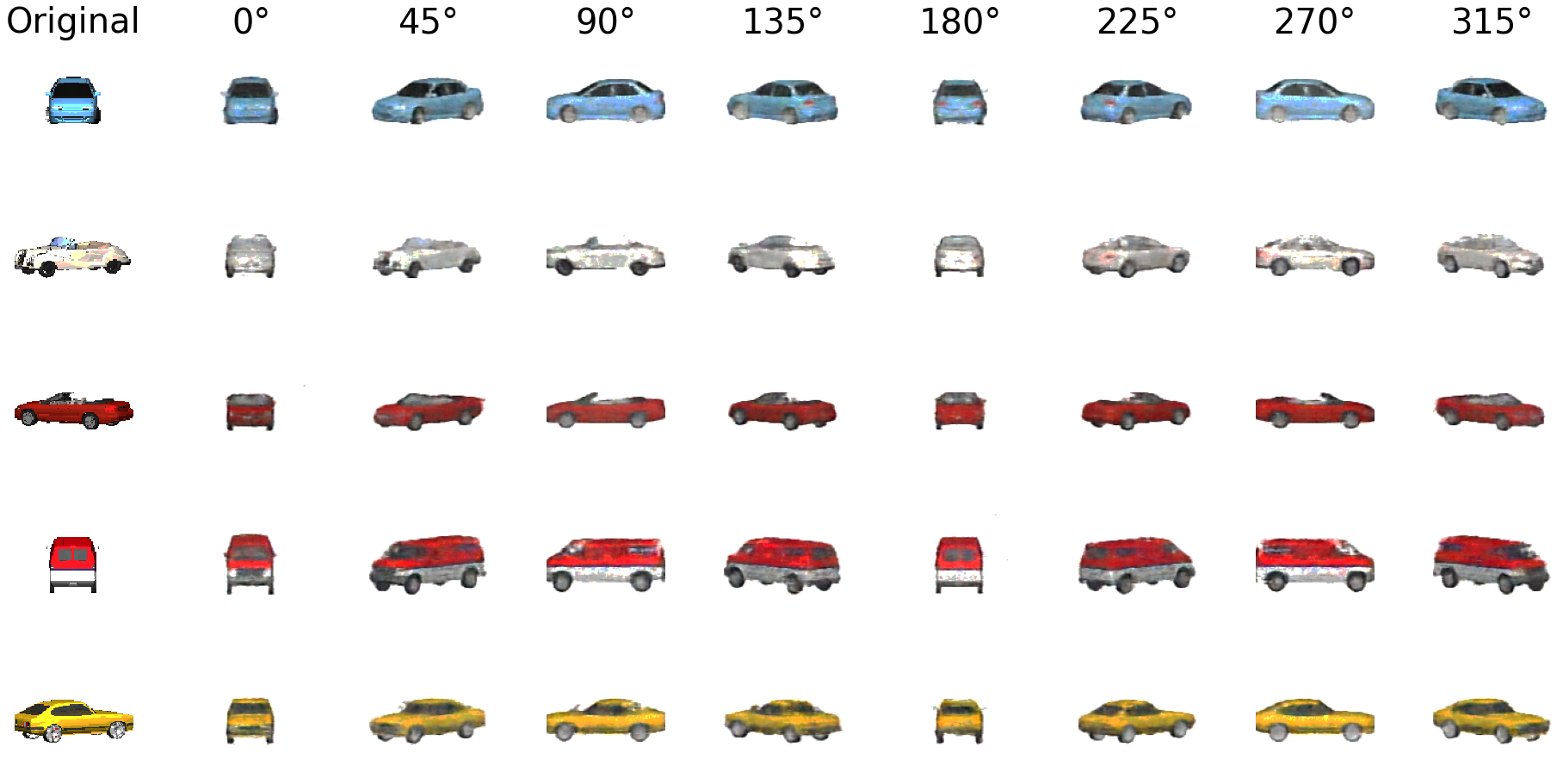}
    \caption{Results for Cars3D}
    \label{fig:cars3d}
\end{figure}

\subsection{Datasets and Experiments Description}
To test the performance of our proposal we used three datasets. The well-known MNIST \cite{MNIST} and SVHN \cite{SVHN} and, in addition, Cars3D \cite{reed2015deep}. The latter contains images of 199 cars rendered at 24 equispaced angles. We randomly chose 796 images for the test and the remaining 3980 for the train. As for the architectures, we use convolutional layers with a kernel size of 3 and stride of 2. In Table \ref{tab:archs} we show the number of channels, layers, and dimensions of the latent variables for each dataset. To train all of them, we used the Adam Optimizer with a batch size of 128 and a learning rate of 0.001 for 100 epochs for MNIST and 500 epochs for SVHN and Cars3D. In all cases, $\beta=0.001$ in the modified ELBO that must be maximized.
\begin{table}[ht]
\caption{Architecure details for each dataset where $d(z_l)$ and $d(z_u)$ are the dimensions for $z_l$ and $z_u$, respectively.}
\label{tab:archs}
\begin{tabularx}{\columnwidth}{cccY}
    \hline
        & $d(z_l)$      & $d(z_u)$      & Channels                      \\ \hline
MNIST   & 512           & 512           & \{16,32,64,128\}              \\ 
SVHN    & 1024          & 1024          & \{32,64,128,256\}             \\
Cars3D  & 1024          & 1024          & \{128,128,256,256,512,512\}   \\ \hline
\end{tabularx}
\end{table}

\vspace{-2mm}
\subsection{Domain Translation Results}
In the following, we show visually the results provided by our method for each of the datasets. In Figure \ref{fig:mnist}, we randomly select some test samples from the MNIST dataset and compare the outcomes generated by our system with those produced by StarGAN \cite{StarGAN2018}, which is the most representative baseline for the task of the present work. In it, we see that the numbers generated by our system are much more realistic, and also that the background is less noisy, as in the original images. Likewise, in Figure \ref{fig:svhn} we show the results for SVHN. In this case, we see that the style and the color of the digit as well as the background are preserved. In addition, in the cases where extra digits appear in the images, these are maintained, modifying only the middle one, which is the one that corresponds to the label during training. Finally, in Figure \ref{fig:cars3d}, we present the outcomes for the Cars3D dataset. Here, we observe that the model and color of the car remain consistent, with only the car's angle being modified. However, in the second-row car, the model fails to generate the correct profile, possibly because it belongs to an uncommon car style in the dataset. For the last two datasets, we refrain from showing results obtained with StarGAN, as the generated images are not realistic.

\subsection{Verifying the Disentanglement}
Finally, to assess if all the relevant information to identify the class is modeled by $z_l$ and if $z_u$ is independent of class information, we build two classifiers, both with a linear layer followed by a softmax. The first one, $C_l$, takes as input $z_l \sim q_\phi(z_l|x)$ and the second one, $C_u$, takes as input $z_u \sim q_\phi(z_u|x)$. Both aim to determine the class to which $x$ belongs. The accuracy of $C_l$ is expected to be very high, whereas that of $C_u$ should be nearly equivalent to random classification. In table \ref{tab:acc} we show the results of the experiment. We see that, indeed, the accuracy is very high in all cases in $C_l$. However, although the accuracy in $C_u$ is very low, it is not equivalent to a completely random classification for the first two datasets (it should be around 10\% in both cases). This may be due either to the fact that in $z_u$ there is still information about the class, or that there is some correlation between class and style in these datasets. In the Cars3D dataset, we have certainty that this correlation does not exist since all the cars appear at all the angles. Here, the results are similar to a random classification. Therefore, we could think that there is some correlation between class and style in the MNIST and SVHN samples, although it is difficult to verify and remains beyond the scope of this paper.

\begin{table}[ht]
\caption{Accuracy (\%) for $C_l$ and $C_u$ in the different datasets}
\label{tab:acc}
\begin{tabularx}{\columnwidth}{cYYY}
    \hline
        & MNIST     & SVHN      & CARS3D    \\ \hline
$C_l$   & 99.29     & 90.78     & 95.58     \\ 
$C_u$   & 25.29     & 19.68     & 3.88      \\ \hline
\end{tabularx}
\end{table}

\section{Conclusion}
\label{sec:conclusion}
In this paper, we have presented a method to solve the unsupervised multiple-domain translation task using exclusively a modified Variational Autoencoder. This model incorporates two distinct latent variables. One is intentionally designed to be influenced by the domain, while the other is meant to capture dependencies related to all factors except the domain. Another peculiarity of our proposal is that, although one of the latent variables depends on the domain, the encoder and decoder are common to all domains. Therefore, we only need to make transformations in the latent variable to perform domain translation. We define these transformations ourselves and impose that they are linear. This, in addition to simplifying the domain translation, allows us to have control and understanding over the latent space. Concretely, we design this latent space such that to carry out the domain translation we must perform a rotation of the domain latent variable, while the other latent variable must remain intact so that properties not associated with the domain are not modified. 

We have verified that our proposal satisfactorily solves the objective task for different datasets outperforming other well-known works which, in addition, have higher complexity. Finally, we have verified that the two latent variables have a high degree of disentanglement, so that the one associated with the domain contains all the relevant information about the domain, while the other one hardly contains any information about the domain.

\clearpage
\bibliographystyle{IEEEbib}
\bibliography{strings}

\end{document}